\pdfoutput=1

\documentclass[11pt]{article}

\usepackage[final]{acl}

\usepackage{times}
\usepackage{latexsym}

\usepackage[T1]{fontenc}

\usepackage[utf8]{inputenc}

\usepackage{microtype}

\usepackage{inconsolata}

\usepackage{graphicx}

\usepackage{booktabs}
\usepackage{multirow}
\usepackage{subcaption}
\usepackage{amsmath}

\newcommand{\todo}[1]{}
\newcommand{\liz}[1]{}
\newcommand{\hannah}[1]{}

\newcommand{\dataset}[1]{\textsc{StoryFeedback} #1}

\title{Help Me Write a Story: \\Evaluating LLMs' Ability to Generate Writing Feedback}
\author{Hannah Rashkin \quad Elizabeth Clark \quad  \textbf{Fantine Huot} \quad \textbf{Mirella Lapata}
\\
 Google DeepMind \\
 \texttt{\{hrashkin, eaclark, fantinehuot, lapata\}@google.com}
}

\begin{document}
\maketitle
\begin{abstract}
Can LLMs provide support to creative writers by giving meaningful writing feedback? 
In this paper, we explore the challenges and limitations of model-generated writing feedback by defining a new task, dataset, and evaluation frameworks. To study model performance in a controlled manner, we present a novel test set of 1,300 stories that we corrupted to intentionally introduce writing issues.  We study the performance of commonly used LLMs in this task with both automatic and human evaluation metrics.  Our analysis shows that current models have strong out-of-the-box behavior in many respects---providing specific and mostly accurate writing feedback.  However, 
models often fail to identify the \emph{biggest} writing issue in the story 
and to correctly decide when to offer critical vs.\ positive feedback.
\end{abstract}

\section{Introduction}
With the growing popularity of LLMs, there is a lot of interest in using these models for human-in-the-loop collaboration such as creative writing tasks where humans may want to use LLMs as writing assistants \citep{Lee_2022,yuan-etal-2022-wordcraft,mirowski-2022-cowritingscreenplays} or educational tasks where LLMs may act as tutors that help human students improve their skills \citep{lee2023learningteachingassistantsprogram,jurenka2024responsibledevelopmentgenerativeai}.  One under-studied task in this area is how well these models can be used as writing tutors that help students by providing feedback on their writing. For instance, discussion from \citet{gero-long-chilton-2023-aisupport} highlights that giving good feedback is an important role in providing AI support to creative writers.

\begin{figure}
    \centering
    \includegraphics[trim={0 3.15in 5.8in 0 }, clip, width=\columnwidth]{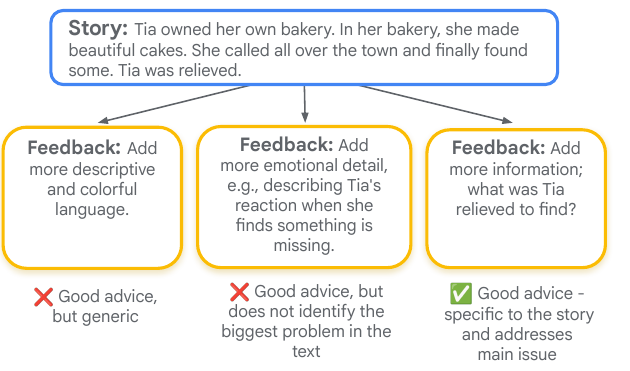}
    \caption{Examples of good and bad feedback for a short story}
    \label{fig:intro}
\end{figure}

Using LLMs to produce writing feedback is a unique use-case that differs from other collaborative writing tasks, such as generating text continuations  \citep[e.g.,][]{Lee_2022} or text rewriting \citep[e.g.,][]{shu-etal-2024-rewritelm}, in that this task doesn't ask the model to do any of the content generation itself. Providing feedback is more open-ended and should help guide the human writer towards fixing the writing themselves. 
Therefore, this task requires a different skill-set from other creative tasks including being able to identify writing problems and articulate them in a clear and constructive way.  However, there aren't many existing resources for studying LLM capabilities in providing feedback outside of academic domains like scientific peer reviews \citep[e.g.,][]{chamoun-etal-2024-automated}.

In this paper, we explore the task of providing open-ended feedback to improve story writing.  Good feedback could include positive comments (e.g., \textit{this writing is great already!}), criticisms (e.g., \textit{the second sentence is confusing}), and/or suggestions for changes (e.g., \textit{give more explanation of this character's motives}).  Beyond just being correct, good feedback should also be specific and well-explained, such as the right-most output in Figure~\ref{fig:intro}.

To explore this task, we create a new dataset consisting of stories, some of which have intentional writing issues.  For this purpose, we design a pipeline for introducing these issues into stories randomly chosen from a seed set of gold story data. We use a mix of the gold stories and corrupted stories as inputs to eight commonly used LLMs, obtaining model-generated feedback for each. The \dataset dataset\footnote{\url{https://github.com/google-deepmind/igen}} includes a total of $\sim$84k pairs of stories with model-generated feedback.

Through in-depth automatic analysis and human annotation, we investigate whether commonly used LLMs are able to provide feedback that: (1) is well-formed and specific, (2) provides correct criticisms and suggestions, (3) correctly identifies the biggest writing problems, especially in the corrupted stories, and (4) gives positive encouragement appropriately.  Our findings demonstrate that most of the models provide well-formed, specific feedback and their suggestions are mostly correct.  However, human scores also show that models struggle more to identify the biggest writing issues, instead commenting on smaller issues.  Our results also show that models face difficulty in deciding when to offer positive comments, often doing so when stories have bigger issues that should be addressed.

\section{Related Work}
\paragraph{Creative Writing Assistance}
There is a large body of work suggesting the utility and drawbacks of using NLP systems as tools for creative writers \cite{chakrabarty2023art,Lee_2022,clark-etal-2018-creativewriting}.
A variety of tasks have emerged for using AI for creative writing, ranging from fully automated story writers \cite{fan-etal-2018-hierarchical,huot2024agents} to more collaborative writing assistants \cite{akoury-etal-2020-storium,Lee_2022,chakrabarty2022help,yuan-etal-2022-wordcraft,Ippolito2022CreativeWW,mirowski-2022-cowritingscreenplays}. There is also a strong interest in aligning these systems with the needs of actual creative writers.   \citet{Guo2024FromPT} underscored the need for LLMs that collaborate in a way that aligns with creative writers' values, particularly the importance of allowing for control and transparency while also being flexible to multiple writing roles.  Recent work studying writers' use of LLMs \cite{gero-long-chilton-2023-aisupport,Chakrabarty2024CreativitySI} has also highlighted the potential of AI-generated feedback to support writers, particularly if the feedback is specific and well-explained.   However, there are scarce resources and evaluations for supporting this task within NLP. In our paper, we work to bridge this gap by providing resources for evaluating the ability of LLMs to generate correct and useful feedback for story drafts.  

\paragraph{Identifying Writing Issues and Revisions} There are multiple prior works \cite{dou-etal-2022-gpt,goyal-etal-2022-falte} presenting methods of identifying writing issues in text.  This is closely related to our task, but those papers focus on identifying categories of writing errors for use in evaluating machine-written text, whereas we focus on describing the writing issues constructively for humans to improve their own writing.
Prior work \cite{du-etal-2022-read,laban-etal-2024-inksync,kim-etal-2022-improving} has also used models to provide direct revisions to a document, automatically or with a human-in-the-loop. Some of this work uses data with edit histories such as Wikipedia articles \cite{du-etal-2022-understanding-iterative,faruqui-etal-2018-wikiatomicedits,shu-etal-2024-rewritelm} that serve as drafts in a similar way to our automatically corrupted versions of the story documents.  In our analysis, we use the edit taxonomies proposed by \citet{du-etal-2022-understanding-iterative} as a guide for the types of writing edits a model might suggest as feedback, though we apply these to story writing rather than Wikipedia editing. 
Rather than using AI to suggest text revisions directly, we propose using the model to provide writing feedback that would help people make their own edits, giving the writer more agency over the editing process.

\paragraph{Feedback and Peer Review}  \citet{benharrak-etal-ondemandfeedback} showed that LLMs are promising tools for providing writers with on-demand, personalizable feedback; however, they also observed that outputs could be wordy, generic, and repetitive.  These observations are mirrored by \citet{han-etal-2024-llm} which explored using essay scores as a way of guiding model output towards more critical, direct essay feedback. In \citet{nair-etal-2024-closing}, they focus on modeling approaches for improving feedback of economic essays with simulated writers.  \citet{yuan-etal-2024-llmcrit} introduces a novel framework for models to use task criteria to create feedback for multiple tasks including scientific writing and reddit story post writing. Complementary to this line of recent work, we contribute new large-scale analysis with human evaluation across multiple dimensions of feedback quality as well as providing new data resources for this area.
Whereas much of the existing work in generating feedback has focused on academic writing domains, such as generating peer reviews for scientific articles \cite{dycke-etal-2023-nlpeer,kang-etal-2018-dataset,chamoun-etal-2024-automated,darcy2024margmultiagentreviewgeneration} or academic essay writing \cite{nair-etal-2024-closing}, we provide data and evaluation resources to explore how LLMs can be applied to generating feedback that suits the needs of the creative writing domain, more specifically.

\begin{figure}
    \centering
    \includegraphics[trim={0 3.2in 5.1in 0},clip,width=.9\columnwidth]{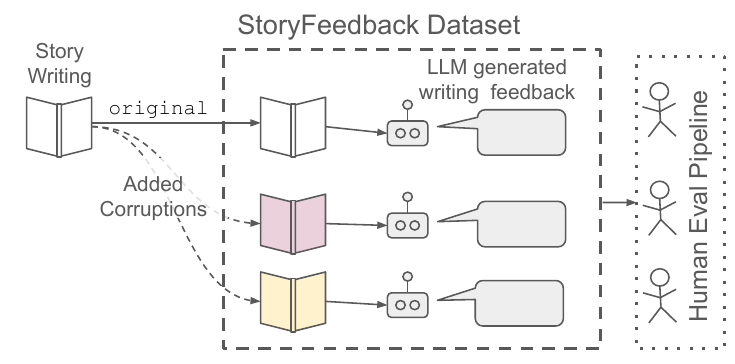}
    \caption{An overview of the task. Short stories are corrupted and given to models to generate feedback, which is then evaluated along several dimensions.}
    \label{fig:task_overview}
\end{figure}

\section{The Task: Story Writing Feedback}
\label{sec:task}
Our goal is to understand how users  
can utilize LLMs as collaborative tools for improving their writing. In this work, we scope the task to asking LLMs to provide story-writing feedback, which we loosely define to include any type of positive encouragement, actionable suggestions, or constructive criticisms.  
We also focus on a single turn interaction (i.e. asking an LLM for feedback on a single writing draft), though this could be extended easily to multiple turns as a draft is edited iteratively by the writer.

We design new data and evaluation frameworks for exploring LLM capabilities at this task, depicted in Figure ~\ref{fig:task_overview}.  For the purpose of large-scale analysis, we focus the scope of this paper's investigation to short stories with controlled writing issues (created heuristically in Section~\ref{sec:dataset}) that can be easily scaled up and curated.  The intention of the analysis in this paper is to act as an initial exploration into the challenges of this task and to develop more formal evaluation criteria that can also be applied to wider scopes of writing domains (such as long-form narratives).

In order to perform this task, the model should ideally give the writer feedback that is specific to their story, addresses the story's biggest issues, and makes the story better. More concretely, while developing the data and evaluation frameworks, we consider the following questions regarding the model output:
\begin{enumerate}
    \item \textbf{Is it valid feedback that's specific to the story?} The output should be well-formed and assess the quality of the story in some way. The feedback should be tailored to the input story, not generic writing advice.
    \item \textbf{Would following the feedback make the story better?} If the writer follows the feedback, it should improve the quality of the story.
    \item \textbf{Does it identify the biggest problem in the story?} The feedback should focus on whatever the main issues are in the text, rather than suggesting minor changes.
    \item \textbf{Is the model giving positive comments appropriately?} It's important for the model to be able to state positive feedback (such as appraising that the story is already high-quality).  But, on the other hand, it shouldn't only give positive feedback if there are still writing issues that need to be addressed.
\end{enumerate}

\begin{figure*}
    \centering
    \includegraphics[trim={0 2.9in 0 0},clip,width=.9\textwidth]{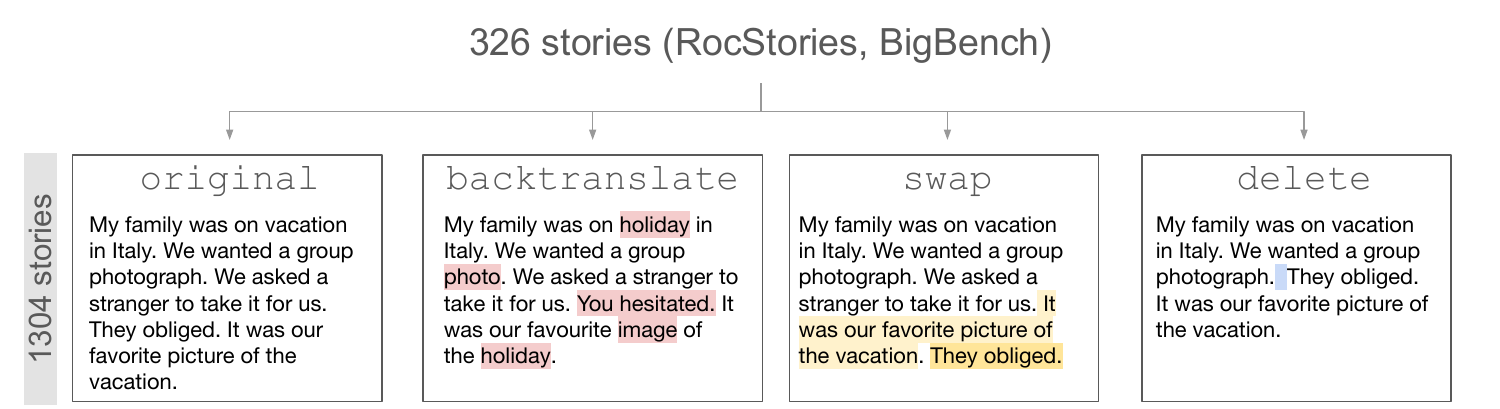}
    \caption{Example with the corruption methods applied.  Highlights indicate the portions of the story that are different from the original.}
    \label{fig:data:ex}
\end{figure*}

\section{The \dataset Dataset}
\label{sec:dataset}
We present \textsc{StoryFeedback}, a dataset of English short stories paired with model-generated feedback to evaluate how accurately models can identify major issues in a text and articulate suggestions for improving the writing. In total, we collected 83,456 story and feedback pairs.

\paragraph{Seed Stories} As a seed set of stories, we collect 326 short stories\footnote{We choose to focus on short stories that can be read and analyzed by human raters more efficiently.} from publicly available datasets: the ROCStories \citep{mostafazadeh-etal-2016-corpus} validation set and three BIG-bench story datasets (Abstract Narratives \cite{ghosh-etal-2022-epic}, English Proverbs, and English Fables) \citep{srivastava2023beyond}.  We randomly sampled up to 100 stories from each (The English Proverbs data only contains 26 stories). 

\paragraph{Story Corruption Methods} We can't control the type of writing issues that may exist in the stories from the seed set, and many of them may not need additional feedback beyond positive comments. To ensure that we have data with specific sets of writing problems, we create several sets of stories with more controlled, automatically induced writing issues.  We use three methods of corrupting the stories from the seed story set, as described below:
\begin{itemize}
    \item \texttt{backtranslate}: The story is translated\footnote{Using the BertSeq2Seq translation model from \citep{rothe-etal-2020-leveraging}.} to German and then back to English ten times to get a newly perturbed story.\footnote{We chose to translate it ten times based on observing the amount of noise it added for a small set of examples.}  This type of permutation sometimes introduces grammar issues and can change the wording or tone.  It can also disrupt the clarity through slight meaning changes or affect the coreference of pronouns.
    \item \texttt{swap}: Two consecutive sentences in the story are swapped.  This type of permutation typically disrupts the coherence of the story as events no longer take place in the correct order.
    \item \texttt{delete}: A randomly selected sentence is deleted from the story. This can cause events to become confusing since they lack prior context that would have come from a previous sentence.
\end{itemize}
In addition, we also include the original stories (\texttt{original}) in the dataset for comparison of the types of feedback given on the original gold stories vs. the corrupted ones.  In total, our dataset consists of $\sim$1,300 stories (326 each of the types \texttt{original}, \texttt{backtranslate}, \texttt{swap}, and \texttt{delete}). In Figure~\ref{fig:data:ex}, we show four stories from the dataset which were all corrupted from the same seed story.  

\paragraph{Models} We prompt eight models to generate feedback for each corrupted and gold story. We selected this list of models to represent several popular, general-purpose, and publicly available models that people could easily access and use out-of-the-box for writing advice, while covering a range of model quality and size:
\begin{itemize}
    \item Bloomz 7B \& 176B \citep{bloomz} An open-source instruction-following model in 2 sizes.
    \item Gemma 9B \& 27B \citep{gemma} Instruction-tuned versions of the open-source Gemma 2 models in 2 sizes.
    \item Gemini 1.5 Flash \& Pro \citep{gemini} Instruction-tuned versions of the Gemini 1.5 model in 2 sizes.
    \item GPT 3.5 \& 4 \citep{gpt4} 2 versions of the GPT model.
\end{itemize}
More details about the models and generation process are in Appendix~\ref{app:models} with examples in Table~\ref{tab:feedback:modelex}.

\paragraph{Prompting Set-ups}
We explore four prompts for eliciting feedback from the models, varying the amount of information given to the model and the amount of feedback requested.  Our goal in studying these different prompts is to understand the sensitivity of the models to the varying amounts of instructions that a non-expert might give.
\begin{itemize}
    \item \texttt{bulleted list} (BL full): Prompts the model for a bulleted list of feedback for the story.  We also include additional context to the model by providing a list of categories of possible writing issues (e.g., fluency, coherence, etc.) that the feedback should address.  The categories of possible writing issues are derived from the text revision taxonomy proposed by \citet{du-etal-2022-understanding-iterative}.
    \item \texttt{bulleted list only} (BL Only): The same prompt as \texttt{bulleted list}, but without the list of writing suggestion categories.
    \item \texttt{one sentence} (1-Sent): Prompts the model for a single sentence of feedback without further instructions.
    \item \texttt{spot the problem} (SpotProb): Frames the task as a challenge, asking the model to find the one problem in the text and describe it in one sentence.
\end{itemize}

All prompts additionally instructed the model to generate the phrase ``The text is perfect as-is.'' if there were no problems in the story.  
The full prompts are listed in App.~\ref{app:prompts}. For each prompt, we experiment with both a zero-shot and two-shot version of the prompt. The two-shot examples are the same for all prompts (one corrupted story and one original story) and can be found in App.~\ref{app:prompts}.
\section{Automatic Evaluation}
\label{sec:autoevals}

We run several automatic metrics on the generated feedback to describe the output and evaluate its quality, and the results are presented in Table \ref{tab:automatic_results}.

Because we intentionally introduced errors into the stories, we can evaluate how often the model fails to identify these errors.
Every prompt instructed the model to say, ``The text is perfect as-is.'' if it did not see any problems in the story.
Ideally, a model should not say a story is perfect if the story has been corrupted by one of our strategies (\texttt{backtranslate}, \texttt{swap}, or \texttt{delete}).
We look at what percentage of the feedback contains the phrase ``is perfect as-is'' (labeled as \textit{\% PAI} in Table \ref{tab:automatic_results}), and if it contains the phrase, how often it is for an \texttt{original} (i.e., uncorrupted) story (\textit{PAI prec}).
More concretely, \begin{equation}
\text{\textit{PAI prec}} = \frac{\text{\# ``perfect as-is'' for \texttt{original} stories}}{\text{\# ``perfect as-is'' for all stories}}
\end{equation}
where \textit{\# ``perfect as-is''} represents the number of generated outputs that contain the phrase ``perfect as-is.''
Ideally, $\text{\textit{PAI prec}}=1$, because the feedback should never say a corrupted story is perfect.
We focus on the precision of ``perfect-as-is''  because the other type of errors (i.e., giving negative feedback to uncorrupted stories) is not as concerning since it is still possible to give good critical feedback to an uncorrupted story.

\begin{table}[t]
    \centering
    \small
    \begin{tabular}{@{}l@{~}l@{~~}c@{~~}c@{~~}c@{~~}c@{~}c@{~}c@{}}
    \toprule
    & & length & \% {PAI} & {PAI} &  3-gramR  & 3-gramR \\
    & & & & prec &  & (w/o {PAI})  \\
    \midrule
    \multirow{8}{*}{\rotatebox[origin=c]{90}{\small models}} & Bloomz 7B & 88.31 & 0.32 & 0.24 & 0.77 & 0.68 \\
    & Bloomz 176B & 124.74 & 0.26 & 0.25 & 0.67 & 0.57 \\
    & Gemma 9B & 343.62 & 0.06 & 0.43 & 0.47 & 0.44 \\
    & Gemma 27B & 354.31 & 0.03 & 0.46 & 0.46 & 0.44 \\
    & Gemini 1.5 Flash & 443.46 & 0.04 & 0.51 & 0.40 & 0.37 \\
    & Gemini 1.5 Pro & 482.10 & 0.07 & \textbf{0.52} & \textbf{0.38} & \textbf{0.34} \\
    & GPT 3.5 & 176.23 & 0.43 & 0.29 & 0.70 & 0.48 \\
    & GPT 4 & 349.08 & 0.25 & 0.40 & 0.55 & 0.41 \\
    \midrule
    \multirow{4}{*}{\rotatebox[origin=c]{90}{\small prompts}} & bulleted list only & 542.62 & 0.15 & 0.31 & 0.51 & 0.43 \\
    & bulleted list & 445.12 & 0.23 & \textbf{0.34} & 0.65 & 0.56 \\
    & spot the problem & 96.62 & 0.20 & 0.31 & \textbf{0.50} & \textbf{0.38} \\
    & one sentence & 96.56 & 0.15 & 0.30 & 0.53 & 0.44 \\
    \midrule
    \multirow{2}{*}{\rotatebox[origin=c]{90}{\small nshot}} & zeroshot & 366.83 & 0.12 & 0.28 & \textbf{0.49} & \textbf{0.42} \\
    & twoshot & 223.63 & 0.24 & \textbf{0.34} & 0.61 & 0.49 \\

    \bottomrule
    \end{tabular}
    \caption{Automatic evaluation results, broken down three different ways: by model, by prompt, and by zeroshot vs. twoshot. \textit{Length} is the number of characters. \textit{PAI} stands for ``perfect-as-is,'' \textit{prec} abbreviates ``precision,'' and \textit{3-gramR} stands for ``trigram repetition.'' All numbers are averages across the entire subset. \textbf{Bold} indicates the best-performing category (high is best for PAI prec and low is best for 3-gramR).}
    \label{tab:automatic_results}
\end{table}

\begin{figure*}[tb!]
    \centering
    \includegraphics[trim={0cm 1cm 0 2.08in},clip,width=\textwidth]{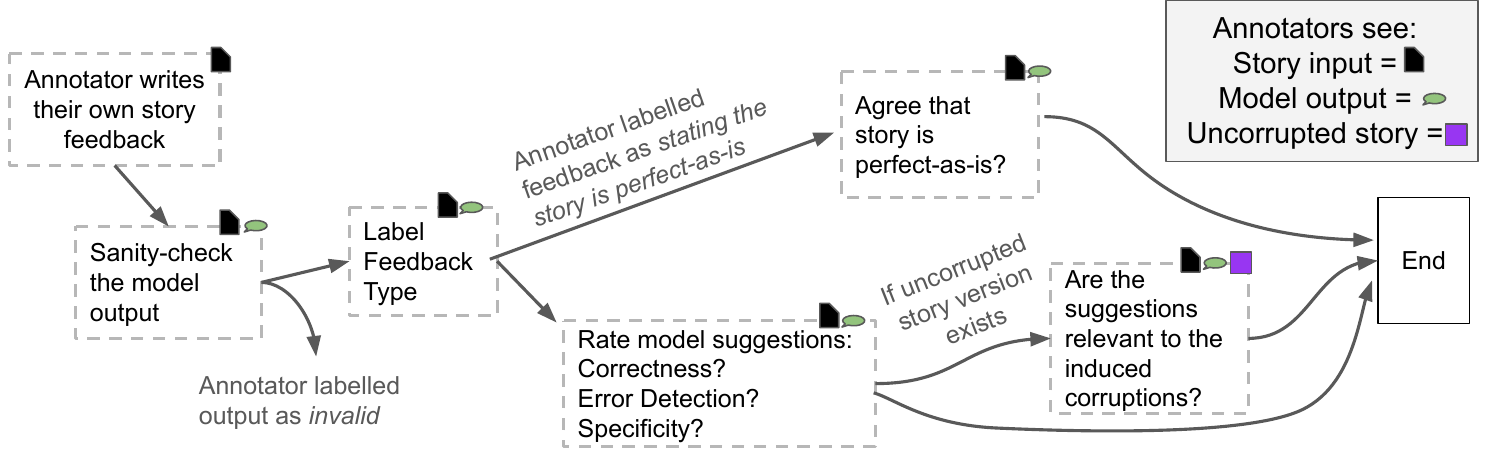}
    \caption{Outline of the human annotation task in which annotators can follow different question paths.}
    \label{fig:ann:task:diagram}
\end{figure*}

Although the precision scores are low across all subsets of the data, the larger versions of the models are more precise in using the ``perfect as-is'' feedback option than their smaller counterparts.
The Gemma and Gemini models are the most precise; however, they also generate this feedback far more infrequently than the other models ($<10\%$ of the time).
The two-shot feedback is more precise than the zero-shot, perhaps because one of the few-shot examples is a case where the text is labeled as perfect as-is.
(The percent of feedback that contains the phrase ``perfect as-is'' doubles between the zero-shot and two-shot feedback.)

We also measure the diversity of the feedback by looking at the trigram repetition\footnote{We used the implementation from \citet{huot2024agents}.} between a piece of feedback and all other feedback in that subset (\textit{3-gramR}).
We use this lexical overlap metric to detect the undesirable behavior of repeating the same generic feedback verbatim, regardless of the input story.
A high repetition score means the feedback for one story frequently overlaps with feedback for other stories (i.e., low diversity).

The larger versions of the models had more diverse outputs than their smaller counterparts, and the Gemma and Gemini models had the most diverse feedback, overall.
However, models that frequently reply with ``the text is perfect as-is'' naturally have worse diversity scores.
To account for this, we also calculate the trigram overlap excluding those pieces of feedback (\textit{3-gramR (w/o PAI)}), and we find that GPT~4 has the most diverse output after the Gemini models.
The bulleted list prompt produces the least diverse output, perhaps because it is the most structured prompt, and models are more likely to copy from the bulleted list structure provided in the prompt. This may also be the reason that providing fewshot examples in the prompt also leads to a drop in the feedback diversity.

Additional automatic analysis and more detailed breakdowns of the results are in App.~\ref{app:automatic_results}.

\section{Human Evaluation}
While automatic evaluations can capture some aspects of the diversity or precision, the goal of the task is to produce useful feedback for writers; with this goal in mind, we run a human evaluation on a subset of the generated feedback.
Participants annotated 1,920 pieces of feedback, randomly selected to be evenly distributed between all the models, corruptions, and prompt types (this comes out to 240 examples annotated per model). We collected 3 human annotations for each piece of feedback.  Annotators responded to questions along multiple dimensions of quality, which were designed to investigate the larger research questions that we posed at the end of Sec.~\ref{sec:task}.

\subsection{Annotation Task}
We show the annotators a story and the corresponding model-generated feedback and ask them to answer a subset of the multiple choice questions below\footnote{Exact question and answer phrasing in App.~\ref{app:humaneval:instructions}}.  Their answers are translated to scalar values between 0 and 1 (details in App.~\ref{sec:app:aggr}). The possible paths through the questions are shown in Figure \ref{fig:ann:task:diagram}.  

\begin{table}[tb]
    \centering
    \begin{tabular}{lcc} \toprule
      Dimensions   & Avg Pair Sim & Kripp.~$\alpha$ \\ 
      \midrule
sanity-check & 0.95& 0.49\\
feedback-type & 0.99& 0.95\\
perfect-agree & 0.58& 0.15\\
correctness & 0.80& 0.40\\
error-detection & 0.68& 0.33\\
specificity & 0.89& 0.61\\
relevance & 0.79& 0.40 \\\bottomrule
    \end{tabular}
    \caption{Inter-annotator agreement measures: Average pairwise similarity (Avg Pair Sim) between pairs of annotator responses on the same example, where similarity is 1 minus the Euclidean distance between responses' numerical values. The Krippendorff's alpha (Kripp.~$\alpha$) was computed using the same Euclidean distance metric between responses.}
    \label{tab:iaa}
\end{table}
\begin{table*}[tb]
    \centering

\resizebox{\textwidth}{!}{
    \begin{tabular}{clccccccc}\toprule
Open-source&Model &	sanity &  feedback-type & perfect-agree &	correctness	 & error-detection & 	specificity &	relevance\\ 
\midrule
\parbox[t]{2mm}{\multirow{2}{*}{\rotatebox[origin=c]{0}{Yes}}} 
& Bloomz 7b	& $0.874\pm0.02$	& $0.410\pm0.03$	& $0.285\pm0.04$	& $0.292\pm0.03$	& $0.219\pm0.03$	& $0.288\pm0.03$	& $0.226\pm0.03$\\
& Bloomz 176b	& $0.792\pm0.02$	& $0.291\pm0.03$	& $0.450\pm0.04$	& $0.316\pm0.02$	& $0.254\pm0.03$	& $0.460\pm0.04$	& $0.243\pm0.02$\\
\midrule

\parbox[t]{2mm}{\multirow{2}{*}{\rotatebox[origin=c]{0}{Yes}}}& Gemma 9b	& $0.993\pm0.0$	& $0.062\pm0.02$	& {$ \bf0.656\pm0.07$}	& $0.734\pm0.02$	& $0.609\pm0.02$	& $0.904\pm0.02$	& $0.462\pm0.03$\\
& Gemma 27b	& {$ \bf 0.997\pm0.0$}	& $0.051\pm0.01$	& $0.485\pm0.07$	& $0.755\pm0.02$	& $0.614\pm0.02$	& $0.902\pm0.02$	& $0.455\pm0.02$\\\midrule

\parbox[t]{2mm}{\multirow{2}{*}{\rotatebox[origin=c]{0}{No}}}& Gemini Flash	& $0.989\pm0.0$	& $0.026\pm0.01$	& $0.367\pm0.11$	& $0.763\pm0.02$	& $0.687\pm0.02$	& $0.917\pm0.01$	& $0.481\pm0.02$\\
& Gemini Pro	& $0.982\pm0.01$	& $0.052\pm0.01$	& $0.625\pm0.10$	& $0.792\pm0.02$	& $0.703\pm0.02$	&{$ \bf0.953\pm0.01$}	& $0.497\pm0.02$\\
\midrule

\parbox[t]{2mm}{\multirow{2}{*}{\rotatebox[origin=c]{0}{No}}}& GPT 3.5	& $0.989\pm0.0$	& {$ \bf0.419\pm0.03$}	& $0.427\pm0.03$	& $0.766\pm0.02$	& $0.663\pm0.03$	& $0.862\pm0.02$	& $0.438\pm0.03$\\
& GPT 4	& $0.989\pm0.0$	& $0.199\pm0.03$	& $0.573\pm0.05$	& {$ \bf0.834\pm0.02$}	& {$ \bf0.757\pm0.02$}	& $0.942\pm0.01$	& {$ \bf0.593\pm0.03$}\\

\bottomrule

    \end{tabular}}
    \caption{The average score and standard error of the mean (SEM) of human responses per dimension per model, aggregated over all other variations of nshot, prompt type, and noise type.  Note: the Gemma/Gemini models rarely say the story is perfect as is, so the perfect-agree scores for these models are computed over very small sample sizes ($n<20$) and have a large SEM margin.}
    \label{tab:human_main_model}
\end{table*}

\begin{enumerate}
    \item Sanity-check: Is the model-generated output phrased as feedback about the story?  The goal of this criterion is to filter out any degenerate or irrelevant outputs.  {\it Options: Yes; No; Contains both}
    \item Feedback-type: Is the feedback positive (e.g., ``The story is perfect as-is.'') or does it provide suggestions. {\it Options: Positive comments; Suggestions; Contains both}
    \item Perfect-agree: If the feedback says that the story is perfect as-is, do you agree? {\it Options: Yes; No}
    \item Correctness: If the feedback makes a suggestion, would following the suggestion improve the story? {\it Options: Yes; Some of the suggestions; No}
    \item Error-detection:  Would following the suggestion fix the \emph{biggest} issue with the story? {\it Options: Yes;  No; I don't see any issues in the story}
    \item Specificity: Is the suggestion specific to this story? {\it Options: Yes; No}
    \item Relevance: If the story was corrupted, would following the feedback fix the corruption?\footnote{For this question, we show annotators the original story and ask whether the feedback reflects the changes between the original and corrupted stories.} {\it Options: Yes; Some of the suggestions; No}
\end{enumerate}

We also ask annotators to provide their own feedback for the story writer at the beginning of the task, prior to seeing the machine-generated feedback. 
This encourages annotators to read the story carefully before answering the questions and also produces a set of human-authored feedback for our analysis.

\subsection{Annotation Statistics}
For the multiple choice questions, we compute the interannotator agreement (Table~\ref{tab:iaa}) using the Euclidean similarity between the values of the responses of two annotations for the same example.  The Krippendorff's alpha scores for most dimensions are moderate to high indicating general agreement.
The dimension with the lowest agreement scores is perfect-agree (whether the annotators agree that the story is ``perfect-as-is'').  We posit that this question has more natural ambiguity since it's subjective whether a text is already ``good enough'' or if more changes can be made.  The error-detection question had higher agreement than the perfect-agree but also indicated some ambiguity.  When computing the evaluation scores, we use the averages of the scores from all three raters per example, as an aggregated response.

More details about the human evaluation implementation (including evaluator training and score aggregation) are in App.~\ref{app:humanevals}.

\subsection{Human Evaluation Results}
In Tables~\ref{tab:human_main_model}--\ref{tab:human_main_prompt_type} we report the average human ratings aggregated by model, noise type, and prompt type.

\paragraph{Difference in Model Architectures} Based on the results in Table~\ref{tab:human_main_model}, we note that most models perform quite highly at the specificity question and also reasonably highly on the correctness question. But, they have weaker performance in the error-detection and relevance dimensions.  It seems the models miss what the annotators felt were the most important issues (error-detection) or what issues were introduced via the corruption strategy  (relevance).  Overall, this indicates that models provide concrete and plausible suggestions but often leave out suggestions that would target the most salient writing problems.  

Additionally, the models tend to score lowly on the perfect-agree question, indicating that the models are often wrong when they say that the story doesn't need any further edits, which is similar to  our observations about the low precision scores from the automatic results (Table~\ref{tab:automatic_results}).  Also mirroring the automatic metrics, we do see that the feedback-type is ``is-perfect-as-is'' in only a very small percentage  of cases from the Gemini and Gemma models (about 5\% of the time), confirming that these model families are more prone towards frequent suggestions.

Performance may be slightly affected by model size, but these  improvements are small and depend on the dimension. 
We observe that the proprietary models outperform the open source models (particularly outperforming over Bloom) across most dimensions.  
The two largest models (Gemini Pro and GPT 4) perform similarly to one another, especially when considering the margin of error, though each class of model may have different strengths across different dimensions. GPT 4 in particular has strong performance in the error-detection and relevance dimensions.

\begin{table*}[tb]
    \centering
\resizebox{\textwidth}{!}{
    \begin{tabular}{lccccccc}\toprule
Corruption Type &	sanity & feedback-type &	perfect-agree &	correctness	 & error-detection & 	specificity &	relevance\\
\midrule
\texttt{original}	& $0.958\pm0.01$	& $0.241\pm0.02$	& $0.581\pm0.03$	& $0.630\pm0.02$	& $0.485\pm0.02$	& $0.825\pm0.02$	& $nan\pm nan$\\
\texttt{backtranslate}	& $0.946\pm0.01$	& $0.124\pm0.02$	& $0.301\pm0.04$	& $0.770\pm0.01$	& $0.712\pm0.02$	& $0.840\pm0.02$	& $0.577\pm0.02$\\
\texttt{delete}	& $0.955\pm0.01$	& $0.213\pm0.02$	& $0.422\pm0.03$	& $0.666\pm0.02$	& $0.545\pm0.02$	& $0.813\pm0.02$	& $0.367\pm0.02$\\
\texttt{swap}	& $0.943\pm0.01$	& $0.158\pm0.02$	& $0.331\pm0.04$	& $0.691\pm0.02$	& $0.621\pm0.02$	& $0.815\pm0.02$	& $0.378\pm0.02$\\
\bottomrule
    \end{tabular}}
    \caption{The average score and standard error of the mean (SEM) of human responses for each type of story input, aggregated over all other variations of n-shot, prompt type and model type.}
    \label{tab:human_main_noise}
\end{table*}
\paragraph{Effects of Story Corruption} As shown in Table~\ref{tab:human_main_noise}, \texttt{backtranslate} has higher scores for correctness, error-detection, and relevance compared to other corruption types, implying models can spot those types of errors more easily.
Writing issues caused by swapping or deleting sentences (e.g., incoherent events and missing information) are not as easily detected.  This is most notable from the relevance category, where models struggle to make suggestions that would be relevant to the swapped or deleted alterations. 
Annotators are more likely to agree with the positive comments on \texttt{original} stories (perfect-agree) than the positive comments on the corrupted ones, overall.  However, when the model tries to provide constructive suggestions to an \texttt{original} story, it is more likely to offer low-quality suggestions (low correctness and error-detection) compared to the suggestions for the corrupted stories.

\begin{table*}[tb]
    \centering
    
\resizebox{\textwidth}{!}{
    \begin{tabular}{lccccccc}\toprule
Prompt &	sanity & feedback-type &	perfect-agree &	correctness	 & error-detection & 	specificity &	relevance\\
\midrule
SpotProb	& $0.894\pm0.01$	& $0.216\pm0.02$	& $0.461\pm0.03$	& $0.577\pm0.02$	& $0.437\pm0.02$	& $0.788\pm0.02$	& $0.414\pm0.02$\\
1-Sent	& $0.962\pm0.01$	& $0.139\pm0.02$	& $0.444\pm0.04$	& $0.689\pm0.02$	& $0.538\pm0.02$	& $0.741\pm0.02$	& $0.439\pm0.02$\\
BL only	& $0.960\pm0.01$	& $0.174\pm0.02$	& $0.446\pm0.04$	& $0.746\pm0.01$	& $0.679\pm0.02$	& $0.923\pm0.01$	& $0.406\pm0.02$\\
BL Full	& $0.986\pm0.0$	& $0.205\pm0.02$	& $0.389\pm0.03$	& $0.747\pm0.01$	& $0.719\pm0.02$	& $0.844\pm0.02$	& $0.513\pm0.02$ \\
\midrule
0-shot	& $0.919\pm0.01$	& $0.152\pm0.01$	& $0.387\pm0.03$	& $0.722\pm0.01$	& $0.628\pm0.01$	& $0.849\pm0.01$	& $0.450\pm0.01$ \\
2-shot	& $0.982\pm0.0$	& $0.206\pm0.01$	& $0.463\pm0.02$	& $0.674\pm0.01$	& $0.577\pm0.01$	& $0.804\pm0.01$	& $0.446\pm0.01$\\\bottomrule
    \end{tabular}}
    \caption{The average score and standard error of the mean (SEM) of human responses for different prompt-types, aggregated.}
    \label{tab:human_main_prompt_type}
\end{table*}
\paragraph{Sensitivity to Prompt Types} As shown in Table~\ref{tab:human_main_prompt_type}, prompting the model to output more information via a bulleted list (BL only and BL full) yields the best results.   Depending on the dimension, it may be more beneficial to provide a taxonomy of possible feedback types (BL full) rather than just requesting a bulleted list (BL only). Prompting the model to ``spot the problem'' in the writing seems to yield the lowest performance in terms of correctness and error-detection.  This is surprising in that it seems that the models struggle to identify the biggest issues when more directly prompted to spot problems in the writing.

\paragraph{Difference in $n$-shot Performance} When comparing the results from zero-shot vs. two-shot models (Table~\ref{tab:human_main_prompt_type}), there is no clear improvement, and it may even harm the scores in certain dimensions to have two-shot set-ups.  Qualitatively, we observe that the two-shot models often become biased towards repeating the feedback from the two-shot examples rather than generating feedback that's more specific to the input story (as reflected in the specificity scores). 

\begin{figure}[tb]
    \centering
    \includegraphics[width=\columnwidth]{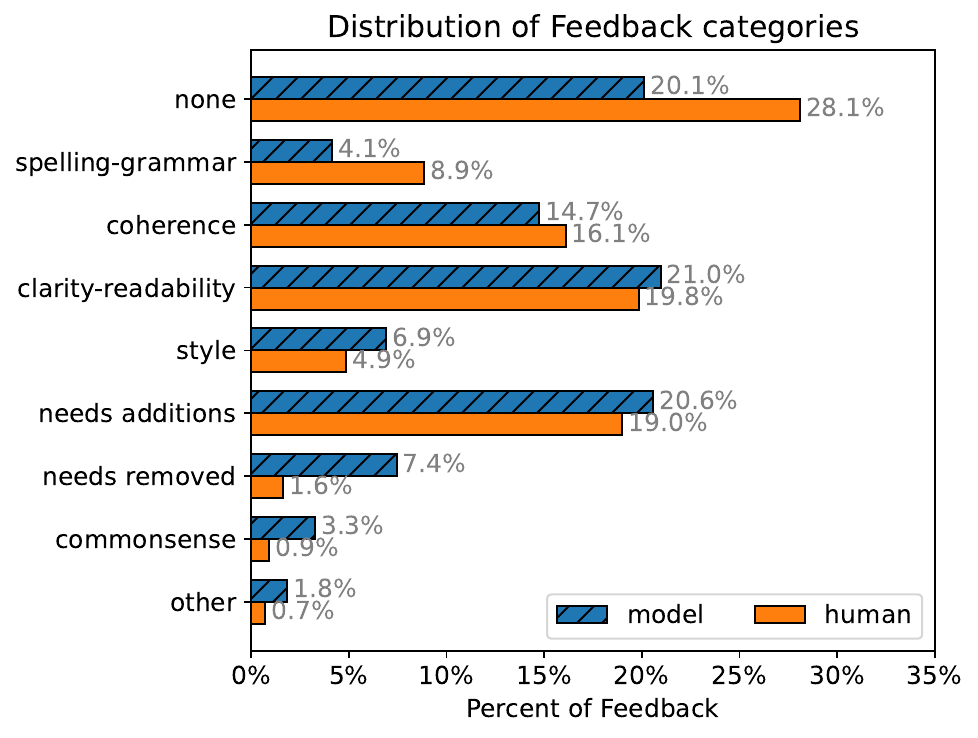}
    \caption{Distribution of categories of feedback written by models vs. humans}
    \label{fig:feedback-cat}
\end{figure}

\subsection{Human-Authored Feedback}
As part of the annotation task, annotators wrote down their own feedback for each story.  We analyze how their feedback compares with the model generations by prompting\footnote{The full prompt is in App.~\ref{app:feedback_prompts}.} Gemini 1.5 Flash to 
categorize\footnote{Based on the categories used in \citet{du-etal-2022-understanding-iterative}.} the types of feedback given by both humans and models and compare the types of issues they mention in Figure~\ref{fig:feedback-cat}.
We find that people are more likely to say there are no issues (especially for the \texttt{original} stories) and give feedback on spelling, grammar, and coherence issues, while model-generated feedback comments more on style and adding or removing information from the story. More details and analysis are in App.~\ref{app:human_feedback}.

\section{Discussion}

We close by revisiting the main questions we posed in Section~\ref{sec:task}.
\paragraph{Can LLMs produce valid feedback that is specific to a story?}  Generally, yes.  Though it may vary by model and prompt type, most of the LLMs reached very high scores in the sanity-check questions indicating that they are typically well-formed valid outputs.  The models generally get high specificity scores as well, meaning that the feedback is typically not just generic comments.
\paragraph {Would following the feedback make the story better?} 
The best models typically achieve correctness scores in the range of 0.7 to 0.8. This still leaves room for improvement, but does indicate that their suggestions are frequently helpful for improving in the story.
\paragraph{Does it identify the biggest problem in the story?} This is still a challenge even for the best performing models. Their error-detection scores are much lower than their correctness scores, which means that, even when they correctly identify issues in the story, they are often missing what annotators' considered the most salient issue.  This is also reflected in the low relevance scores which show that the model suggestions often don't address the intentional errors that we've added by corrupting the stories.
\paragraph{Is the model giving positive comments appropriately?} This is another challenge for models. Both our automatic and human evaluations indicate that models are not very discerning with positive encouragement.  When the models call stories ``perfect,'' people frequently disagreed, stating that those stories still contained writing problems that should have been addressed.

\paragraph{Implications for future work} In this paper, we focus solely on short stories with synthetic writing issues to allow for systematic analysis that could be scaled over many data examples and model architectures.  The analysis on these stories is useful as a first step towards uncovering challenges in this task; however, to understand the full scope of LLM abilities as writing tutors, this work would need to be extended to additional domains, such as longer and more complex stories with naturally occurring writing issues. Models may especially struggle to detect and offer feedback for some types of errors in long texts (e.g., contemporaneous work from \citet{ahuja2025findingflawedfictionsevaluating} demonstrates that models struggle to identify continuity errors in longer stories).
In more realistic writing settings, human writers may also use a model for feedback over multiple iterations of draft-writing. 
While we only consider a single turn of feedback in our setting, it remains an open research question whether these LLMs would be consistent and useful in settings where a writer may repeatedly ask for feedback over multiple draft iterations. Collecting resources towards these domains can be challenging as human annotations for longer stories and iterative writing sessions also requires significantly more work from human annotators.  As a first step in this direction, our paper provides \dataset{} as a resource to facilitate future work and our discussion of the research questions above as starting points for further investigation.

\section{Conclusion}
We introduce the task of providing writing feedback on short stories, which is challenging for LLMs because it requires both identifying writing issues and formulating constructive advice.  We create the \dataset dataset to explore how well LLMs can generate feedback on simple stories, many of which have induced writing errors.  With in-depth automatic and human evaluations, we uncover some of the strengths and weaknesses of out-of-the-box models for this task.  While models show promising abilities to provide plausible, specific, and accurate feedback, they sometimes miss the more salient problems in favor of lower hanging fruit.  We suggest that future work in this area should focus on designing models that are able to identify the most salient writing issues rather than just providing ``correct'' feedback as well as extending these evaluation frameworks to long-form narratives with more complex writing issues.

\section*{Limitations}
Our study is limited to model behavior in the domain of English stories, and we choose to focus on relatively short stories that can be analyzed in more depth.  We also synthetically introduce errors into the text to control the type and frequency of writing errors. Observations may differ on longer, more complex stories, for non-English languages, or for naturally occurring writing errors.  Additionally, we limit our study to eight common LLMs with four different prompts, to characterize a range of performance behaviors, but do not make an exhaustive search of all possible models or prompt choices.  We also choose to focus on specific dimensions of feedback quality like the correctness, relevance, specificity, etc., but there may be other qualities like the tone of the feedback that should be explored in future work. Lastly, we acknowledge that there is some subjectivity in what makes feedback ``useful,'' and as noted in our discussion of the human annotations, there may be variations in judgment due to the natural ambiguity of the task. 

\section*{Ethical Considerations}
In designing LLMs that will provide support to human writers, it's important to consider how this might align with writers' values \cite{Guo2024FromPT}, and, namely, integrating these systems in a way that does not diminish the writers' agency, control, and transparency in their own creative decisions.  A challenge of using these systems with students, in specific, is the concern that they may use this technology as a crutch rather than further honing their skills. In this task, an LLM may provide feedback that has a negative impact on their writing (for example, making suggestions to include things that involve stereotypes or plagiarism) or their engagement (for example, providing overly harsh critical feedback). Future work with more focus on investigating these issues is important for understanding how these systems could be used responsibly in downstream applications.

\section*{Acknowledgments}
We would like to thank the anonymous reviewers for their insightful comments and suggestions.  We would also like to thank Annie Louis, Edward Grefenstette, Sumit Asthana, Tuhin Chakrabarty, Chaitanya Malaviya, Chris Alberti, Shereen Ashraf, Jennimaria Palomaki, Shoshana Jakobovits, Ankur Parikh, Dipanjan Das, Slav Petrov, Chris Dyer, and our other Google Deepmind collaborators for providing constructive feedback and computational resources at multiple stages of this project. Lastly, we also thank the annotators who participated in our human evaluation studies.

\bibliography{custom, anthology}

\appendix

\section{Model Details}
\label{app:models}
The Bloom feedback was generated from checkpoints downloaded from Hugging Face\footnote{\url{https://huggingface.co}} in May 2024. The GPT model output was generated using the OpenAI API\footnote{\url{https://platform.openai.com/}} in May 2024. The Gemma and Gemini models were generated from the publicly available checkpoints\footnote{\url{https://aistudio.google.com/}} in December 2024.

The feedback was generated using the API or model's default settings to mimic scenarios where writers would use these systems out-of-the-box. All model outputs were sampled with temperature $=0.70$.

\section{Prompts}\label{app:prompts}
The full text for each of the four prompt types and the twoshot examples are as follows:

\subsection{Bulleted List}

\noindent Write a bulleted list of feedback for the following text:

\noindent <STORY>

\noindent The feedback should state whether the text is:

- not fluent (e.g. typos, grammar issues)

- not coherent (lacking internally consistency or badly ordered)

- lacking clarity (e.g. text is wordy or ambiguous)

- needs a stylistic change (e.g. formality, tone, other writing preferences)

- needs additions  (e.g. needs more information or updated information)

- needs something removed  (e.g. get rid of something that doesn't fit with the rest of the document)

\noindent If the story has no problems, say `The text is perfect as-is'.

\noindent Feedback:

\subsection{Bulleted List Only}
\noindent Write a bulleted list of feedback for the following text:

\noindent <STORY>

\noindent If the story has no problems, say `The text is perfect as-is'.

\noindent Feedback:

\subsection{One Sentence}
\noindent Write one sentence of feedback for improving the following text:

\noindent<STORY>

\noindent The feedback should be brief and address the most important problem in the text.

\noindent If the story has no problems, say `The text is perfect as-is'.

\noindent Feedback:

\subsection{Spot the Problem}
\noindent There is one problem in this text. Can you find it?

\noindent <STORY>

\noindent In one sentence, describe the problem.

\noindent If the story has no problems, say `The text is perfect as-is'.

\noindent Feedback:

\subsection{Twoshot Examples}
\noindent \textbf{Example 1}

\noindent Story: Cathy loves basketball. She made plans to go watch her favorite team play. So she went to the store to buy more. She sat in the front row. Her team won the game.

\noindent Feedback:\footnote{\label{footnote:feedback_format}The feedback is preceded by a dash (- ) for the \texttt{bulleted list} and \texttt{bulleted list only} prompts.} The third sentence in the text is incoherent with the rest of the story and should be removed.

\noindent \textbf{Example 2}

\noindent Story: Matty's cats kept scratching themselves. She didn't know what to do. Eventually she found out they had fleas. She gave them flea medicine. They were completely healed after that.

\noindent Feedback:\footnote{See footnote \ref{footnote:feedback_format}.} The text is perfect as-is.

\section{Additional Automatic Evaluation Results}\label{app:automatic_results}
This section contains several additional automatic evaluations of the generated feedback.
\begin{table}[t]
    \centering
    \small
    \begin{tabular}{lccc}
    \toprule
    & & trigram rep & BERTScore  \\
    \midrule
    \multirow{8}{*}{\rotatebox[origin=c]{90}{\small models}} & Bloomz 7B & 0.10 & 83.74 \\
    & Bloomz 176B & 0.11 & 83.98 \\
    & Gemma 9B & 0.09 & 83.89 \\
    & Gemma 27B & 0.10 & 84.00 \\
    & Gemini 1.5 Flash & 0.10 & \textbf{84.17} \\
    & Gemini 1.5 Pro & \textbf{0.11} & 84.05 \\
    & GPT 3.5 & 0.05 & 83.67 \\
    & GPT 4 & 0.08 & 84.09 \\
    \midrule
    \multirow{4}{*}{\rotatebox[origin=c]{90}{\small prompts}} & bulleted list only & 0.08 & 84.01 \\
    & bulleted list & 0.05 & 83.12 \\
    & spot the problem & \textbf{0.15} & \textbf{84.37} \\
    & one sentence & 0.09 & 84.28 \\
    \midrule
    \multirow{2}{*}{\rotatebox[origin=c]{90}{\small nshot}} & zeroshot & \textbf{0.12} & \textbf{84.14} \\
    & twoshot & 0.07 & 83.76 \\
    \bottomrule
    \end{tabular}
    \caption{The similarity between a story and its feedback measured by the average trigram overlap (trigram rep) and BERTScore. The results are broken down three different ways: by model, by prompt, and by zeroshot vs. twoshot. Higher similarity scores indicate more overlap with the story, and the highest numbers are in \textbf{bold}. All numbers are multiplied by 100 for readability.}
    \label{tab:story_trigram_repetition}
\end{table}
\subsection{Trigram Overlap with Story}
In Section \ref{sec:autoevals} we present the trigram overlap between one piece of feedback and the rest of the feedback in a subset to understand the diversity of a subset.
We can also look at the trigram overlap between a piece of feedback and the story it is based on as a measure of the specificity of the feedback. Higher overlap may indicate that the feedback is referencing specific portions of the story in its feedback.
Table \ref{tab:story_trigram_repetition} shows the average trigram overlap between the story and the feedback, aggregated across different splits of the data.
It also shows the average BERTScore \citep{bert-score} for each subset (an embedding-based similarity metric).

\subsection{Corruption Method}
\begin{table}[]
\centering
    \begin{tabular}{lr}
    \toprule
    noise & \% \textit{perfect as-is} \\
    \midrule
    original & 0.23 \\
    backtranslate & 0.13 \\
    random sentence swap & 0.15 \\
    delete one sentence & 0.21 \\
    \bottomrule
    \end{tabular}
    \caption{The percent of feedback that included the phrase ``perfect as-is'' broken down by the type of noise that was introduced into the story.}
    \label{tab:noise_results}
\end{table}
The frequency of the ``perfect as-is'' feedback also provides insight into which story corruption methods were the most challenging for the models to detect.
As shown in Table \ref{tab:noise_results}, the models were most likely to say a corrupted story was perfect when the corruption method was \texttt{delete}.
They did so at a rate close to the original (i.e., uncorrupted) stories (0.21 vs. 0.23), indicating the deleted sentences were less likely to be caught by the models than backtranslation or sentence swapping.
\subsection{One Sentence Instruction Following}
\begin{table}[]
\centering
\small
    \begin{tabular}{@{}l@{~}r@{~~}r@{~~}r@{~~}r@{~~}}
    \toprule
    & \multicolumn{2}{c}{one sentence} & \multicolumn{2}{c}{spot the problem} \\
    & zeroshot & twoshot & zeroshot & twoshot \\
    \midrule
    Bloomz 7B & 0.62 & 0.90 & 0.78 & 0.95 \\
    Bloomz 176B & 0.64 & 0.96 & 0.95 & 0.29 \\
    Gemini 1.5 Flash & 1.00 & 1.00 & 0.99 & 0.96 \\
    Gemini 1.5 Pro & 0.99 & 0.99 & 0.99 & 0.97 \\
    Gemma 9B & 0.99 & 0.98 & 0.92 & 0.83 \\
    Gemma 27B & 1.00 & 0.99 & 0.99 & 0.99 \\
    GPT 3.5 & 1.00 & 1.00 & 0.95 & 0.99 \\
    GPT 4 & 0.98 & 0.99 & 0.99 & 0.98 \\
    \bottomrule
    \end{tabular}
    \caption{The percent of feedback that was only one sentence, broken down by prompt type.}
    \label{tab:one_sentence_results}
\end{table}
Two of the prompts (one sentence and spot the problem) instruct the model to output one sentence of feedback.
We evaluate how often the feedback follows these instructions by using the sentence splitting implementation from \citet{huot2024agents} and counting percentage of model responses that only contain one sentence (Table \ref{tab:one_sentence_results}).
The models are generally able to follow this instruction very well.
The Bloom models struggle in some of the zeroshot cases but improve in the twoshot setting.
The exception is the Bloomz 176B consistently output an extra ``Feedback: ''  when given the twoshot spot the problem prompt, which was split into multiple sentences by the splitter.

\subsection{Full Automatic Evaluation Results}
\begin{table*}[]
    \centering
    \tiny
    \begin{tabular}{p{0.25cm}p{0.25cm}p{1.5cm}p{1.25cm}p{2cm}p{2cm}p{2.2cm}p{2.5cm}}
    \toprule
     & & {\small model} & {\small length} & {\small \% \textit{perfect as-is}} & {\small\textit{perfect as-is} precision} & {\small trigram repetition} & {\small trigram repetition (w/o \textit{perfect as-is})} \\
\midrule
    \multirow{16}{*}{\rotatebox[origin=c]{90}{\small bulleted list}} & \multirow{8}{*}{\rotatebox[origin=c]{90}{\small zeroshot}} & Bloomz 7B & 75.51 & 0.27 & 0.24 & 0.88 & 0.85 \\
& & Bloomz 176B & 96.66 & 0.06 & 0.28 & 0.93 & 0.93 \\
& & Gemma 9B & 625.96 & 0.06 & 0.48 & 0.56 & 0.54 \\
& & Gemma 27B & 602.12 & 0.02 & 0.44 & 0.55 & 0.54 \\
& & Gemini 1.5 Flash & 980.99 & 0.02 & 0.45 & 0.47 & 0.47 \\
& & Gemini 1.5 Pro & 1011.8 & 0 & 0.75 & 0.43 & 0.43 \\
& & GPT 3.5 & 323.63 & 0.57 & 0.31 & 0.76 & 0.49 \\
& & GPT 4 & 838.02 & 0.1 & 0.42 & 0.59 & 0.57 \\
    \cmidrule{2-8}
    & \multirow{8}{*}{\rotatebox[origin=c]{90}{\small twoshot}} & Bloomz 7B & 72.15 & 0.69 & 0.24 & 0.92 & 0.8 \\
& & Bloomz 176B & 122.63 & 0.41 & 0.24 & 0.77 & 0.65 \\
& & Gemma 9B & 420.65 & 0.22 & 0.42 & 0.62 & 0.52 \\
& & Gemma 27B & 454.35 & 0.12 & 0.47 & 0.58 & 0.52 \\
& & Gemini 1.5 Flash & 465.07 & 0.16 & 0.48 & 0.52 & 0.44 \\
& & Gemini 1.5 Pro & 457.32 & 0.21 & 0.52 & 0.51 & 0.4 \\
& & GPT 3.5 & 162.69 & 0.49 & 0.35 & 0.76 & 0.52 \\
& & GPT 4 & 412.38 & 0.28 & 0.47 & 0.59 & 0.44 \\
    \midrule
    \multirow{16}{*}{\rotatebox[origin=c]{90}{\small bulleted list only}} & \multirow{8}{*}{\rotatebox[origin=c]{90}{\small zeroshot}} & Bloomz 7B & 71.54 & 0.51 & 0.25 & 0.78 & 0.58 \\
& & Bloomz 176B & 178.72 & 0.14 & 0.29 & 0.19 & 0.08 \\
& & Gemma 9B & 833.17 & 0 & 0 & 0.43 & 0.43 \\
& & Gemma 27B & 836.19 & 0.01 & 0.2 & 0.41 & 0.41 \\
& & Gemini 1.5 Flash & 1134.44 & 0 & 0 & 0.35 & 0.35 \\
& & Gemini 1.5 Pro & 1298.6 & 0.01 & 0.15 & 0.34 & 0.34 \\
& & GPT 3.5 & 466.04 & 0.17 & 0.28 & 0.53 & 0.44 \\
& & GPT 4 & 844.28 & 0.1 & 0.37 & 0.47 & 0.43 \\
    \cmidrule{2-8}
    & \multirow{8}{*}{\rotatebox[origin=c]{90}{\small twoshot}} & Bloomz 7B & 197.42 & 0.23 & 0.21 & 0.95 & 0.94 \\
& & Bloomz 176B & 165.59 & 0.25 & 0.26 & 0.71 & 0.67 \\
& & Gemma 9B & 462.98 & 0.08 & 0.41 & 0.46 & 0.42 \\
& & Gemma 27B & 549.64 & 0.02 & 0.42 & 0.44 & 0.43 \\
& & Gemini 1.5 Flash & 547.54 & 0.05 & 0.51 & 0.42 & 0.39 \\
& & Gemini 1.5 Pro & 661.52 & 0.08 & 0.52 & 0.4 & 0.35 \\
& & GPT 3.5 & 172.5 & 0.36 & 0.3 & 0.67 & 0.49 \\
& & GPT 4 & 261.74 & 0.46 & 0.39 & 0.66 & 0.37 \\
\midrule
    \multirow{16}{*}{\rotatebox[origin=c]{90}{\small one sentence}} & \multirow{8}{*}{\rotatebox[origin=c]{90}{\small zeroshot}} & Bloomz 7B & 88.38 & 0.04 & 0.34 & 0.37 & 0.34 \\
& & Bloomz 176B & 98.06 & 0.35 & 0.25 & 0.47 & 0.21 \\
& & Gemma 9B & 101.03 & 0 & NaN & 0.44 & 0.44 \\
& & Gemma 27B & 90.67 & 0 & NaN & 0.49 & 0.49 \\
& & Gemini 1.5 Flash & 99.29 & 0 & NaN & 0.37 & 0.37 \\
& & Gemini 1.5 Pro & 118.86 & 0 & NaN & 0.35 & 0.35 \\
& & GPT 3.5 & 90.46 & 0.33 & 0.23 & 0.66 & 0.49 \\
& & GPT 4 & 150.59 & 0.04 & 0.36 & 0.44 & 0.42 \\
    \cmidrule{2-8}
    & \multirow{8}{*}{\rotatebox[origin=c]{90}{\small twoshot}} & Bloomz 7B & 59.62 & 0.45 & 0.24 & 0.85 & 0.74 \\
& & Bloomz 176B & 61.37 & 0.48 & 0.24 & 0.92 & 0.84 \\
& & Gemma 9B & 94.67 & 0.04 & 0.48 & 0.49 & 0.47 \\
& & Gemma 27B & 100.64 & 0 & 0 & 0.42 & 0.42 \\
& & Gemini 1.5 Flash & 102.02 & 0.03 & 0.65 & 0.48 & 0.47 \\
& & Gemini 1.5 Pro & 100.48 & 0.08 & 0.53 & 0.44 & 0.39 \\
& & GPT 3.5 & 93.78 & 0.2 & 0.39 & 0.63 & 0.53 \\
& & GPT 4 & 95.09 & 0.34 & 0.42 & 0.6 & 0.4 \\
\midrule
    \multirow{16}{*}{\rotatebox[origin=c]{90}{\small spot the problem}} & \multirow{8}{*}{\rotatebox[origin=c]{90}{\small zeroshot}} & Bloomz 7B & 72.21 & 0.09 & 0.32 & 0.47 & 0.41 \\
& & Bloomz 176B & 16.78 & 0.01 & 0.21 & 0.45 & 0.41 \\
& & Gemma 9B & 108.45 & 0.01 & 0.36 & 0.33 & 0.32 \\
& & Gemma 27B & 110.5 & 0 & 0 & 0.38 & 0.38 \\
& & Gemini 1.5 Flash & 110.21 & 0 & 0 & 0.2 & 0.2 \\
& & Gemini 1.5 Pro & 110.37 & 0 & 0.5 & 0.21 & 0.21 \\
& & GPT 3.5 & 37.94 & 0.85 & 0.26 & 0.91 & 0.42 \\
& & GPT 4 & 117.17 & 0.16 & 0.38 & 0.38 & 0.26 \\
    \cmidrule{2-8}
    & \multirow{8}{*}{\rotatebox[origin=c]{90}{\small twoshot}} & Bloomz 7B & 69.63 & 0.27 & 0.24 & 0.95 & 0.94 \\
& & Bloomz 176B & 258.1 & 0.4 & 0.22 & 0.79 & 0.79 \\
& & Gemma 9B & 102.08 & 0.07 & 0.41 & 0.45 & 0.41 \\
& & Gemma 27B & 90.36 & 0.11 & 0.49 & 0.41 & 0.34 \\
& & Gemini 1.5 Flash & 108.14 & 0.07 & 0.53 & 0.36 & 0.31 \\
& & Gemini 1.5 Pro & 97.86 & 0.13 & 0.53 & 0.38 & 0.28 \\
& & GPT 3.5 & 62.8 & 0.47 & 0.27 & 0.69 & 0.42 \\
& & GPT 4 & 73.35 & 0.5 & 0.38 & 0.67 & 0.33 \\
\bottomrule
    \end{tabular}
    \caption{Automatic evaluation of feedback, broken down by prompt type, by model, and by zeroshot vs. twoshot prompting.}
    \label{tab:full_automatic_results}
\end{table*}

The full breakdown of automatic evaluation scores are in Table \ref{tab:full_automatic_results}.

\section{Human Evaluation Details}
\label{app:humanevals}
\subsection{Annotator Details and Task Statistics}
We ran our task with responses from 64 annotators, each annotating 90 story-feedback pairs on average.  The median time spent annotating each example was 82.3 seconds. 
For the free-response feedback written by annotators, we collected a total of 5,754 feedback entries. 
The human-written feedback is on average 56.5~characters long. The annotators had the option to write "n/a" as their feedback if they didn't see any writing issues. Prior to completing the task, annotators completed a tutorial and consent form detailing the study in advance. Annotators are all fluent English speakers, who were paid fairly, above minimum wage for their respective locations.

\subsection{Annotation Instructions}
\label{app:humaneval:instructions}
Below we include the exact phrasing of the instructions, question, and answer options, verbatim to how they were provided to the annotators.  In the actual task we also included examples that they could refer to at any time.

\begin{enumerate}
    
    \item \textbf{Well-formed feedback}\\ \textit{Task Instructions}\\ This question verifies that the feedback contains a critique of the story and stays on-topic. This isn’t asking about the quality of the feedback, just verifying that it is actually feedback for the given story, since we don’t want to evaluate it if it is not. Feedback that starts talking about another story or that rambles about irrelevant information should be marked as ``No.''  \\
     \textit{Q/A Format} \\ Is this feedback for a story, without changing to unrelated topics or adding new stories or  feedback for additional texts? \\
   $\circ$  ``No, the text does not give any feedback.'' \\ 
   $\circ$  ``No. The text includes feedback, but then adds additional off-topic information, stories, or unrequested feedback.''\\
   $\circ$  ``Yes, this is feedback for a story.''\\
    \item \textbf{Feedback Type}\\ \textit{Task Instructions}\\ Does the feedback claim that the story is ``perfect as-is'' or that it doesn't require any changes? If the feedback is simply, ``The text is perfect as-is.'', please label it ``Yes.'' However, if the feedback says the text is perfect, but then also includes other suggestions, please respond ``Both.'' \\
     \textit{Q/A Format} \\ Is the feedback saying that the story is already ``perfect as is'' or is it suggesting further edits: \\
        $\circ$ ``Yes, the only feedback is that the text is perfect as-is.'' \\
        $\circ$ ``No, the feedback does not say the text is perfect and suggests  changes to the story.''\\
        $\circ$ ``Both. It says the text is perfect, but also includes other feedback.''\\
    
    \item \textbf{Perfect-Agree}\\ \textit{Task Instructions}\\ If the feedback doesn't suggest any changes or says the story is ``perfect as-is'', do you agree? If you don’t see any problems or issues with the story, please mark ``Yes.'' If you think there are issues with the story that need to be improved, say ``No.'' Note that this may be subjective in some cases, so please use your best judgment. \\
    \textit{Q/A Format}\\ Do you agree that the story is already good as is?\\
   $\circ$ ``Yes, the story seems fine and doesn't have any major issues.'' \\
  $\circ$ ``No, there are edits I would recommend to improve the story.''\\
    
    \item \textbf{Correctness}\\ \textit{Task Instructions}\\ Does the feedback address weaknesses in the text? It's ok if it doesn't address all the problems in the text, as long as all the things it mentions would improve the story. \\
     \textit{Q/A Format} \\ Correctness: Does the feedback address a weakness in the story?\\
       $\circ$ ``Every piece of feedback is correct.''\\
       $\circ$ ``Some of the feedback is correct, but some of the feedback is not accurate to the text.'' \\
       $\circ$ ``No part of the feedback is correct.'',\\
        
    \item \textbf{Error-Detection}\\ \textit{Task Instructions}\\ If you feel there is a problem with the story, would following the feedback help fix that specific problem? Does any part of the feedback address what you consider to be the biggest issue in the text? As long as one part of the feedback addresses the problem, you can ignore the rest of the feedback. This may be somewhat subjective, so use your best judgment. \\
     \textit{Q/A Format}\\  Error detection:  Does the feedback capture the biggest  problem (if any) with the story?\\
        $\circ$ ``I don’t see any big problems in the story.''\\
        $\circ$ ``Yes, at least one piece of the feedback would help fix the biggest problem with the story.''\\
        $\circ$ ``No, following the feedback would not help fix the biggest problem in the story.''\\

    \item \textbf{Specificity}\\ \textit{Task Instructions}\\ Is it clear that the feedback is for this specific story? Or is it vague or general enough to apply to another story? Specific feedback might mention characters, events, or details from the story, or quote or rewrite pieces of the original story. For this question, it doesn't matter whether the feedback is accurate or helpful, just whether it is specific to the story. \\
   \textit{Q/A Format}  \\ Specificity: Is the feedback specific to this story or could the feedback be applied to other stories? \\
    $\circ$ ``Yes, the feedback is about this story.''\\
    $\circ$ ``No, the feedback is vague enough that it could be about another story.''\\

        \item \textbf{Relevance}\\ \textit{Task Instructions} \\When you read the edited story draft, do the changes that were made from the original story reflect the suggestions from the feedback? Or do the edits and the feedback seem unrelated? If it is a mix of both (i.e., some feedback is reflected in the edits and some is not OR some edits are related to the feedback and others aren't), please select ``Somewhat.'' \\
    \textit{Q/A Format}  \\  Are any of the changes in the edited story relevant to what was suggested in the feedback?\\
    $\circ$ ``Yes, the changes in the story agree with the suggestions in the feedback.''\\
    $\circ$ ``Somewhat, some of the changes in the story reflect some of the  suggestions in the feedback.''\\
    $\circ$ ``No, the changes in the story are unrelated to what the feedback suggested''\\
\end{enumerate}

\subsection{Evaluator Training}
We found that annotators needed training specifically for this task, so we first made them complete a tutorial in which they had to read in-depth task instructions and examples.  Afterwards, they had to pass a qualifying quiz on three randomly chosen stories from a set that was annotated by experts.  For further quality control, we ran analysis on the each annotator after every batch  (average inter-annotator agreement rates, the median time spent per story, the variance in the time they spent per story) and performed manual grading on the annotators with outlier characteristics.  If an annotator did not pass the regular quality control checks, we removed their annotations and re-annotated their responses with different annotators (about 360 responses were re-annotated). 

\subsection{Aggregating Evaluator Scores}
\label{sec:app:aggr}
To aggregate scores, we treat each of the multiple-choice questions as values between $[0,1]$ and then we average the answers from the annotators. For the binary (yes/no) questions, 0 is ``no'' and 1 is ``yes''.  For the questions with a ``somewhat'' category (i.e. relevance \& correctness), the ``somewhat'' is mapped to $0.5$.  For the error-detection question, we translate both of the non-``Yes'' answers (the ``I don't see any issues'' answer and the ``No there would still be an issue'') to 0.  The full phrasing of questions, answers and their mapping to numerical values can be found in Table~\ref{tab:humanevalqtypes}.  Because of the pipelined annotation approach, not every dimension was answered for each example, only the applicable ones.  If the example has two or more values for a given dimension,  we compute the average of the non-missing responses to get the final value.

\section{Validating Automatic Measures}\label{app:validating_automatic_measures} We validate the performance of the automatic metrics from Section~\ref{sec:autoevals} by comparing them to the human evaluation results. For the automatic metrics, we searched for the phrase ``is perfect as-is'' to automatically detect the feedback type; in the human evaluation, we asked evaluators to label whether the text only contained positive feedback. We find a Pearson correlation coefficient of $\sim$0.90, showing a strong correlation between the presence of the phrase ``perfect as-is'' and positive feedback. This indicates that models generally followed the prompt instruction to use the phrase ``The text is perfect as-is.'' when they saw no issues in the story.

We also look at the relationship between the trigram-based diversity metric and the human evaluation specificity ratings. The Pearson correlation of coefficient between non-unique trigrams and the specificity of the response is $-0.55$, indicating a moderate anti-correlation.

\section{Examples of Feedback}
Examples of the model-generated feedback are in Table \ref{tab:feedback:modelex}.

\section{Analysis of Human-Authored Feedback}\label{app:human_feedback}
This section contains further details and discussion of the analysis of the human-authored feedback.

\subsection{Feedback Analysis Prompts}\label{app:feedback_prompts}
The prompt given to Gemini 1.5 Flash to label the feedback:

\noindent\textbf{Feedback Labeling Prompt}

\noindent Please select the category that best summarizes the issue described in the feedback below.
Do not include any other information.

\noindent Categories: none, spelling and grammar, coherence, clarity and readability, style, needs additions, needs removed, commonsense, other.

\noindent Feedback: Sneaked should be snuck

\noindent Category: spelling and grammar

\noindent Feedback: The story doesn't make sense as Einstein created the theory of relativity so why would he be interested and research something he already knows.

\noindent Category: commonsense

\noindent Feedback: The dialogue feels unnatural and overly formal for a simple interaction between a man and a fly.  The fly's speech, in particular, is too sophisticated.

\noindent Category: style

\noindent Feedback: <FEEDBACK>

\noindent Category: 

\subsection{Automatic Labeling Validation}
We evaluate the model-generated labels by comparing them to a set of gold labels; the model-generated label was also in the gold label set 84\% of the time, out of 480 pieces of feedback annotated by the first author.

\subsection{Distribution of Types of Feedback}
\begin{figure*}[t!]
    \centering
    \begin{subfigure}[t]{\columnwidth}
    \includegraphics[width=\columnwidth]{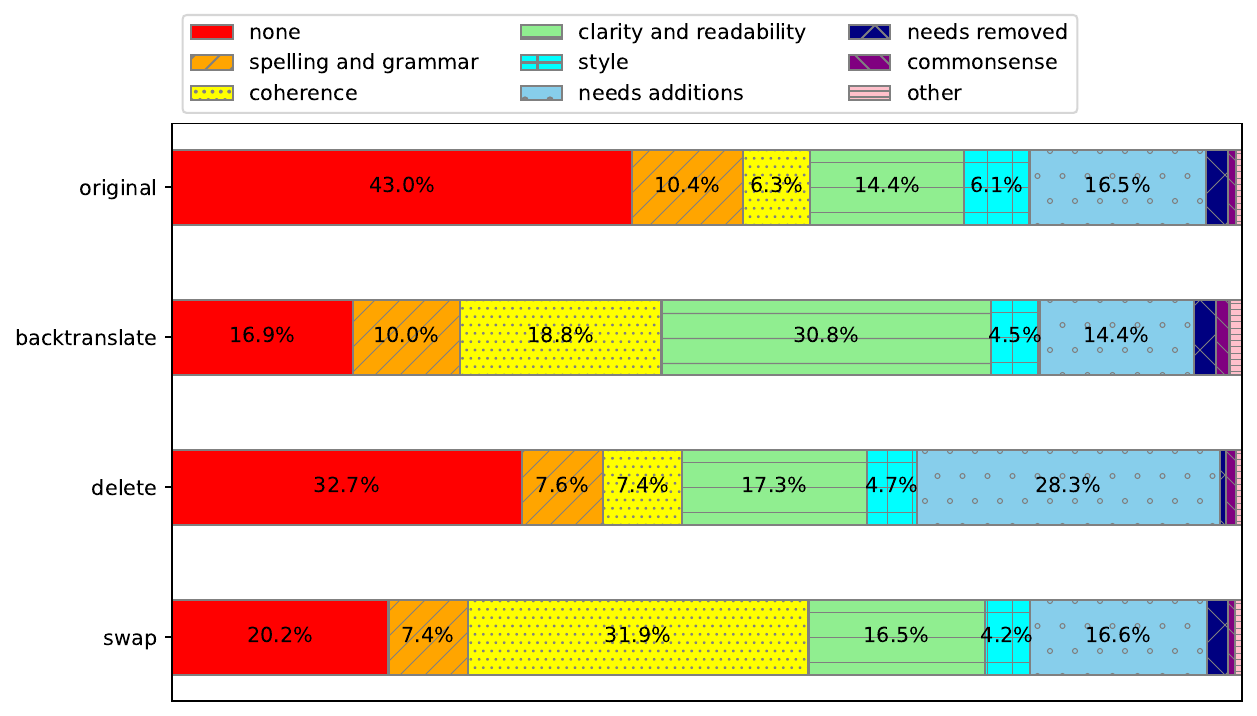}
    \caption{Human Feedback Categories: The categories of feedback provided by human annotators in their free-text responses, split by type of story.  Categories were assigned automatically by Gemini.}
    \label{fig:humanfeedbacktypes}
    \end{subfigure}~~
    \begin{subfigure}[t]{\columnwidth}
    \includegraphics[width=\columnwidth]{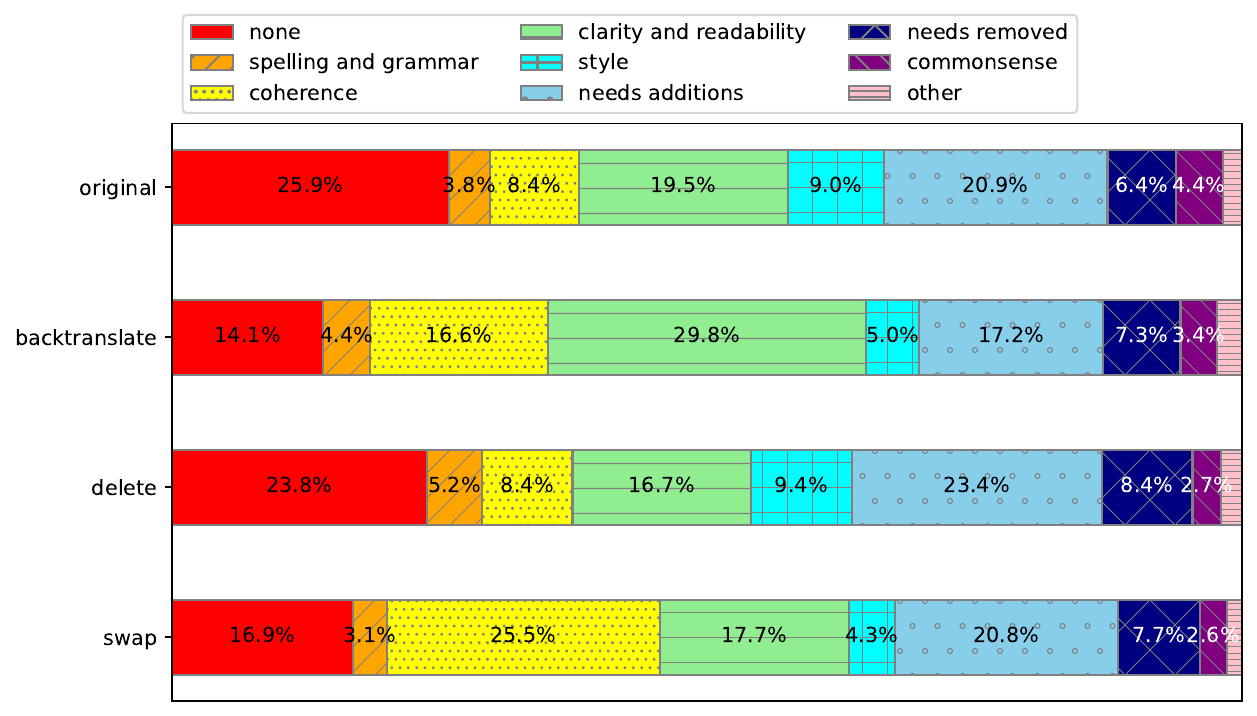}
    \caption{Model Feedback Categories: The categories of feedback provided by models in their output responses, split by type of story. Categories were assigned automatically by Gemini.}
    \label{fig:modelfeedbacktypes}
    \end{subfigure}
    \caption{Types of feedback written by humans and generated by models.}
\end{figure*}

Figure \ref{fig:feedback-cat} shows an overview of the frequency of each label type for model- vs. human-authored feedback.
Figures~\ref{fig:humanfeedbacktypes} and \ref{fig:modelfeedbacktypes} further break down these results by corruption type.

In Figure~\ref{fig:humanfeedbacktypes}, we show the distribution of feedback categories (according to Gemini) of the types of feedback found in human-written feedback collected by annotators.  General trends follow our expectations based on the type of noise that's been added.  Humans give no feedback more often on the original stories.  When reading the \texttt{backtranslate} stories, they generally offer more comments on clarity and readability issues.  When reading \texttt{delete} stories, they offer more suggestions for details that need to be added into the story.  When reading the \texttt{swap} stories, they are most likely to comment on the incoherence of the sequence of events.

In Figure~\ref{fig:modelfeedbacktypes}, we show the distribution of feedback categories of the model-generated feedback. The distributions are mostly similar to the distribution of human-written feedback.  However, the model gives feedback more often (uses the `none' category much less).  Models are also more likely to say things need to be removed.  Unlike humans, models also give feedback criticizing the commonsense of the story, often misunderstanding the intent of the story (e.g., models sometimes gave feedback calling stories unrealistic due to containing talking animals).  Models are less likely than humans to comment on fluency issues (spelling/grammar) overall.

\begin{table*}[]
    \centering
    \footnotesize
    \begin{tabular}{p{.45\textwidth}p{.45\textwidth}}
    \toprule
    \multicolumn{2}{c}{Model-Generated Feedback Examples}\\
     Story   & Feedback\\\midrule
She was unsure as to whether she should open the door or not. Slowly she creaked open the door and peered outside. Her brother leaped out at her and gave her a good scare. She screamed and pushed her brother away. & - The text is fluent with no typos or grammar issues.\newline- The text is coherent, it maintains consistency and is well-ordered.\newline- The text lacks a bit of clarity, especially in the description of the brother's action. It could be more effective to describe exactly how the brother `leaped out'.\newline- The text could benefit from a stylistic change, specifically in varying sentence structure for better flow and engagement.\newline- The text could use some additions, specifically more details to give the reader a better sense of the situation and the characters. For example, the reason why the character was unsure about opening the door could be better explained.\newline- There's no need to remove anything from the text as every part contributes to the overall narrative.\newline- The text could use a bit more emotional description to enhance the reader's connection with the characters. For instance, how the character felt after her brother scared her could be expanded upon. \\\midrule
A rooster found a shiny gem while pecking around in the dirt to find food for himself and his hens. I, however, have no use for you and would have preferred to find some corn, instead. " Looking down at the stone, he observed: " Had your owner found you, he would have picked you up and treasured you.&The text switches from the rooster's perspective to the narrator's perspective inconsistently and abruptly.\\\midrule
A raccoon accidentally stumbled into the theatre's collapse. On his journey he was suddenly blinded by a face that looked at him from above. Impressed by her beauty, he soon realized that it was just a mask worn by actors during their performances, and said, ``Although you are beautiful, it's a shame that you don't have brains.''   & - The first sentence is ambiguous and lacks clarity. It's unclear what ``the theatre's collapse'' refers to. \newline- The order of events could be rearranged for better coherence. For example, the raccoon's realization that the face is a mask could come before his reaction to it. \newline- The story could benefit from additional information or details to provide context. For instance, how did the raccoon end up in the theatre?
\\\midrule
A bat fell to the ground where he was caught by a passing lynx. The bat begged the lynx to spare his life, but the lynx refused, saying that, by her nature, she is the enemy of all birds. Some time later, the bat again fell to the ground and was caught by yet another lynx, whom he also asked to spare him. The lynx said that she is the enemy of all mice, and will therefore kill the bat. The bat, in turn, assured her that he was not a mouse, but a bird. Thus, he was able to escape once again. & The text lacks consistency in its portrayal of the lynx's nature, as it contradicts itself regarding its prey.\\
\bottomrule
    \end{tabular}
    \caption{Examples of Feedback Written by LLMs}
    \label{tab:feedback:modelex}
\end{table*}

\begin{table*}[]
    \centering
    \begin{tabular}{p{5.5cm}|p{6cm}c}
      Question   & Answer Options &\\
\midrule
\multirow{3}{5.5cm}{\parbox{5.5cm}{ [Sanity-Check] Well-formed feedback:  Is this feedback for a story, without changing to unrelated topics or adding new stories or feedback for additional texts?}} & Yes, this is feedback for a story.&$\rightarrow1$\\

&No. The text includes feedback, but then adds additional off-topic information, stories, or unrequested feedback.&$\rightarrow1$\\

&No, the text does not give any feedback.&$\rightarrow0$\\
\midrule
\multirow{3}{5.5cm}{ \parbox{5.5cm}{ [Feedback-type] Is the feedback saying that the story is already “perfect as is” or is it suggesting further edits:}} &Yes, the only feedback is that the text is perfect as-is.& $\rightarrow1$ \\

&Both. It says the text is perfect, but also includes other feedback.&$\rightarrow0$\\

&No, the feedback does not say the text is perfect and suggests changes to the story.& $\rightarrow0$\\

\midrule
\multirow{2}{5.5cm}{\parbox{5.5cm}{[Perfect-Agree] Do you agree that the story is already good as is:}} & Yes, the story seems fine and doesn’t have any major issues.& $\rightarrow1$\\

&No, there are edits I would recommend to improve the story.& $\rightarrow0$\\

\midrule
\multirow{3}{5.5cm}{\parbox{5.5cm}{[Correct] Correctness: Does the feedback address a weakness in the story?}} & Every piece of feedback is correct.& $\rightarrow1$\\
&Some of the feedback is correct, but some of the feedback is not accurate to the text.& $\rightarrow.5$\\
&No part of the feedback is correct.&$\rightarrow0$\\

\midrule
\multirow{3}{5.5cm}{\parbox{5.5cm}{ [Error-Detection] Error detection:  Does the feedback capture the biggest problem [if any] with the story?}} &Yes, at least one piece of the feedback would help fix the biggest problem with the story.&$\rightarrow1$\\

&No, following the feedback would not help fix the biggest problem in the story.& $\rightarrow0$\\

& I don’t see any big problems in the story.&$\rightarrow0$\\

\midrule
\multirow{3}{5.5cm}{\parbox{5.5cm}{[Specificity] Specificity:  Is the feedback specific to this story or could the feedback be applied to other stories?}} & Yes, the feedback is about this story.&$\rightarrow1$\\

&No, the feedback is vague enough that it could be about another story.&$\rightarrow0$\\\\

\midrule
\multirow{3}{5.5cm}{\parbox{5.5cm}{[Relevance] Are any of the changes in the edited story relevant to what was suggested in the feedback? }}& Yes, the changes in the story agree with the suggestions in the feedback.&$\rightarrow1$\\

&Somewhat, some of the changes in the story reflect some of the suggestions in the feedback.&$\rightarrow0$.5\\

&No, the changes in the story are unrelated to what the feedback suggested&$\rightarrow0$\\
\bottomrule
    \end{tabular}
    \caption{Human evaluation questions, full wording and mapping to numerical values}
    \label{tab:humanevalqtypes}
\end{table*}
\end{document}